\documentclass[final]{cvpr}

\usepackage{times}
\usepackage{epsfig}
\usepackage{graphicx}
\usepackage{amsmath}
\usepackage{amssymb}

\usepackage{caption}
\usepackage{subcaption}
\usepackage{amsthm}
\theoremstyle{definition}
\newtheorem{definition}{Definition}[section]
\usepackage{color,soul}
\usepackage{algorithm}
\usepackage{algorithmic}
\newtheorem{theorem}{Theorem}[section]
\newtheorem{corollary}{Corollary}[theorem]
\usepackage{multirow}

\usepackage[pagebackref=true,breaklinks=true,colorlinks,bookmarks=false]{hyperref}

\def\cvprPaperID{6513} 
\def\confYear{CVPR 2021}

\begin{document}

\title{Consensus Maximisation Using Influences of Monotone Boolean Functions}

\author{Ruwan Tennakoon$^1$, David Suter$^2$, Erchuan Zhang$^2$, Tat-Jun Chin$^3$ and Alireza Bab-Hadiashar$^1$\\
$^1$RMIT University, Melbourne Australia. $^2$Edith Cowen University, Perth Australia.\\
$^3$University of Adelaide, Adelaide Australia.\\
{\tt\small \{ruwan.tennakoon, abh\}@rmit.edu.au, \{d.suter, erchuan.zhang\}@ecu.edu.au,}\\ {\tt\small tat-jun.chin@adelaide.edu.au}
%
%
}

\maketitle

\begin{abstract}
Consensus maximisation (MaxCon), which is widely used for robust fitting in computer vision, aims to find the largest subset of data that fits the model within some tolerance level. 
In this paper, we outline the connection between MaxCon problem and the abstract problem of finding the maximum upper zero of a Monotone Boolean Function (MBF) defined over the Boolean Cube. Then, we link the concept of influences (in a MBF) to the concept of outlier (in MaxCon) and show that influences of points belonging to the largest structure in data would generally be smaller under certain conditions. Based on this observation, we present an iterative algorithm to perform consensus maximisation. Results for both synthetic and real visual data experiments show that the MBF based algorithm is capable of generating a near optimal solution relatively quickly. This is particularly important where there are large number of outliers (gross or pseudo) in the observed data. 
   
\end{abstract}

\section{Introduction}

The popular Maximum Consensus (MaxCon) criterion for robust fitting (as typified by that of RANSAC \cite{Fischler_1981}), seeks the maximum sized \textit{feasible} set. Here feasible means that all data points belonging to a ``structure'' (the inlier set) fits its model within a tolerance level. 

Given a set of $n$ data points $\mathcal{D} = \left \{ \mathbf{p}_i\right\}_{i=1}^n$ and a tolerance level $\epsilon$, the MaxCon criterion for robust fitting can be written as:
\begin{align*}
& \underset{\theta,~\mathcal{I}\subseteq \mathcal{D} }{\max} \left | \mathcal{I} \right |\\
& \textrm{subject to} ~~ r_{\mathbf{p}_i}  \left(  \theta \right ) \leq \epsilon \quad \forall \mathbf{p}_i \in \mathcal{I}
\end{align*}
where $r_{\mathbf{p}_i}  \left(  \theta \right )$ is the distance of $\mathbf{p}_i$ from the model $\theta$. 

A subset $\mathcal{I}$ can be represented by length-$n$ bit-vector, $\mathbf{x}$, where the $i$'th position of the bit vector denote the inclusion ($x_i = 1$) or exclusion ($x_i = 0$) of the data point $i$. The above shows that, each subset can be represented by a vertex of the $n$-dimensional Boolean Cube. Therefore, any statement about which of the subsets are feasible,
is a statement on the evaluation of a Boolean function over the $n$-dimensional Boolean Cube. This Boolean function $f : \left \{ 0, 1\right \}^n \rightarrow \left \{0, 1 \right \}$, outputs 1 (for infeasible), or 0 (for feasible) for any vertex of the Boolean Cube. Since MaxCon is a search for the maximum sized feasible subset of the data, it is inherently a search on the Boolean Cube.

Such a view immediately opens the huge theory and very many associated mathematical tools, developed during the study of Boolean functions; for the purposes of the analysis of the MaxCon problem and for devising algorithms to solve this problem. 

Also, there is an additional significant observation. The Boolean function associated with a MaxCon problem belongs to a special class of Boolean functions called the Monotone Boolean Functions (MBF) \cite{Korshunov_2003}.
\theoremstyle{definition}
\begin{definition}
A MBF of $n$ variables is a mapping $f : \left \{ 0, 1\right \}^n \rightarrow \left \{0, 1 \right \}$ such that $\alpha \prec \beta$ implies that $f\left( \alpha \right) \leq f\left( \beta \right)$. Here we apply the natural ordering relation on $n$-dimensional Boolean Cube: $\beta$ follow $\alpha$ ($\alpha \prec \beta$) if for any $i$, the equality $\alpha_i \leqslant \beta_i$ is satisfied.
\end{definition}
This is easy to see: If a subset is feasible, then adding more data to that subset can only move the function towards infeasibility and once infeasible adding more points will not change the subset back to feasible. Likewise, deleting points from a subset can move only towards feasible.  
MaxCon, when viewed in the above sketched framework, is nothing more than the search for the \textit{maximum upper zero} (see the definition below) of the above-mentioned monotone infeasibility function. 
\theoremstyle{definition}
\begin{definition}[Upper zero of a MBF]
Upper zero of the MBF $f\left( \cdot \right)$ is a vertex $\alpha$ for which $f\left( \alpha \right) = 0$ and, for all $\alpha \prec \beta$ the relation $f\left( \beta \right) = 1$ is satisfied. 
\end{definition}

\theoremstyle{definition}
\begin{definition}[Maximum Upper zero of a MBF \cite{KULYANOV_1975}]
Maximum Upper zero of the MBF $f\left( \cdot \right)$ is a vertex $\alpha$ for which $f\left( \alpha \right) = 0$ and, for all $\left \| \beta \right \|_1 > \left \| \alpha \right \|_1$ the relation $f\left( \beta \right) = 1$ is satisfied. Here, $\left \| \alpha \right \|_1 = \sum_i \alpha_i$. 
\end{definition}

\begin{figure}
    \centering
    \begin{subfigure}[b]{0.5\textwidth}
         \centering
         \includegraphics[width=.3\textwidth]{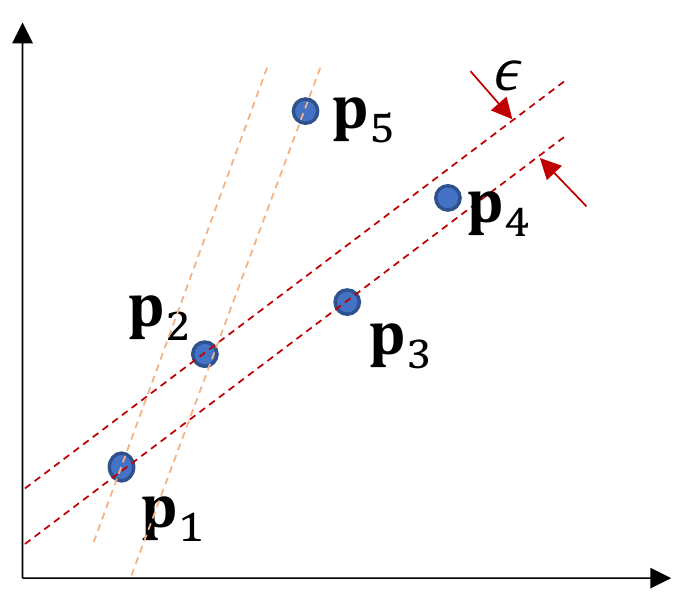}
        \caption{2D line fitting problem}
         \label{fig:bmf_example1}
     \end{subfigure}
     
    \begin{subfigure}[b]{0.5\textwidth}
         \centering
         \includegraphics[width=.9\textwidth]{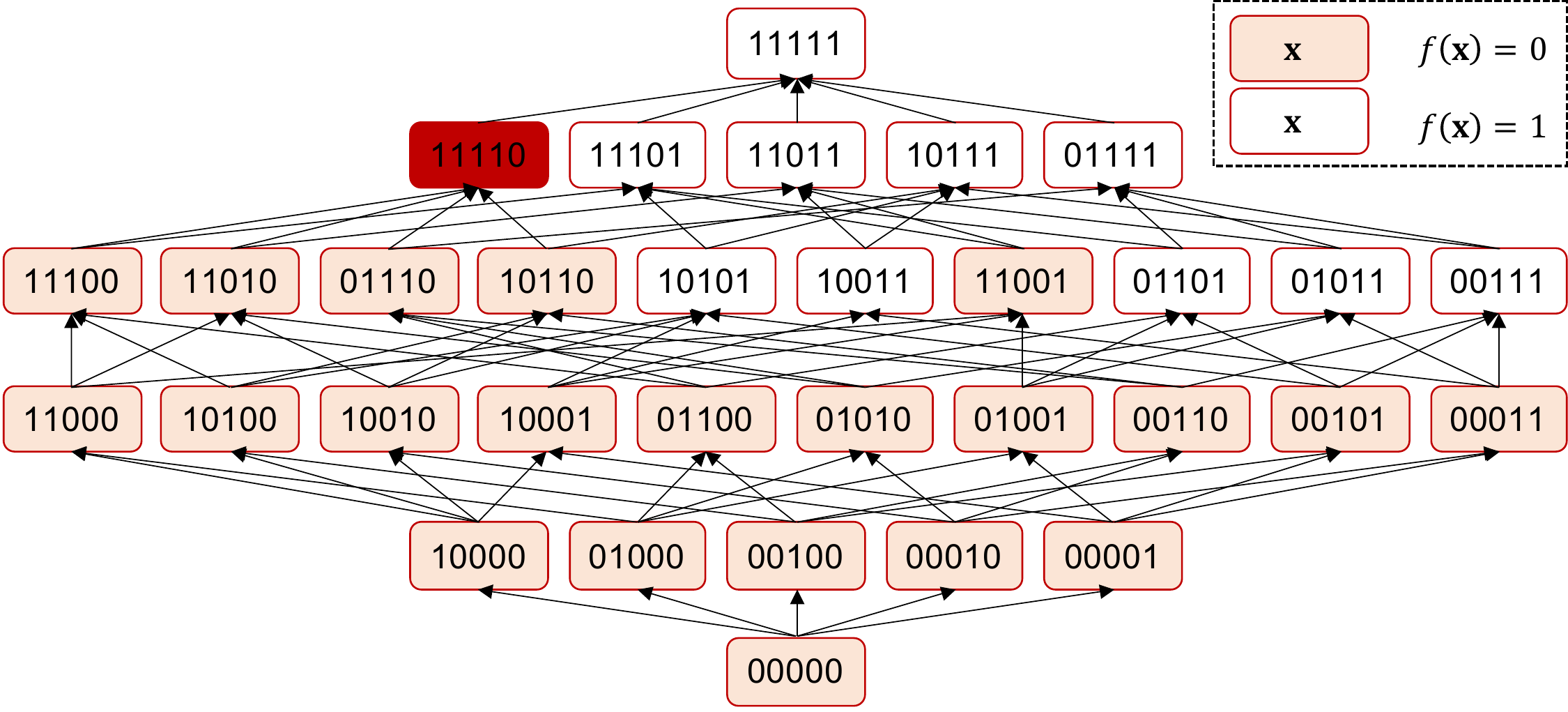}
        \caption{Hasse Diagram}
         \label{fig:bmf_example2}
     \end{subfigure}

    \caption{(a) An example of a 2D line fitting problem with 5 data points together with (b) the associated Boolean Cube and MBF - represented in Hasse diagram format. The 5D Boolean Cube is drawn flattened onto 2D and oriented so that the higher up a vertex appears, the larger is the number of 1's in the coordinates. 
    A Boolean Function maps the Boolean cube to 0 or 1. We illustrate by colouring: ``white'' nodes map to 1, coloured nodes map to 0. This example is a \textit{Monotone} Boolean Function moving up the picture, the value of the function only ever increases, it \textit{never} translate in the opposite direction. Red node is the maximum upper zero in this example.}
    \label{fig:bmf_example}
\end{figure}

A simple example in Figure \ref{fig:bmf_example} illustrates the concepts involved. 
The figure shows that, along paths going up the Hasse diagram (a directed version of the Boolean cube), the function can only stay constant or increase (never decrease). The vertex ``11110'' (representing the subset $\left \{ \mathbf{p}_1, \mathbf{p}_2, \mathbf{p}_3, \mathbf{p}_4 \right \}$) is the MaxCon solution for this problem as it is the highest feasible subset in the Hasse diagram (maximum upper zero). The vertex ``11001'' is an example of an upper zero as there are no feasible nodes that follows it ($\succ \textrm{`11001'}$) on the Hasse diagram (can be seen as a local optimum).    

Whilst Monotone Boolean Functions (MBFs) have been extensively studied for a variety of application domains,
including learning theory \cite{bshouty1996fourier} (and computer vision is, these days, highly dominated by learning style approaches),
there appears to be relatively little attention to MBFs in computer vision: and more specifically in model-based computer vision. Reference \cite{RAMALINGAM_2017} appears to be a recent exception - but even this is tackling  very different considerations from our main areas of interest (namely, \cite{RAMALINGAM_2017} is concerned with efficient Conditional Random Field (CRF) calculations, and CRF modelling is very different to the geometric modelling we refer to).

In this paper, we present a novel view point on the MaxCon problem using the Monotone Boolean Functions theory and use that to develop some efficient solutions. For this purpose, we concentrate on a property of all Boolean Functions, called Influence \cite{o_donnell_2014} (definition in Section \ref{sec:Fourier}); a property that has a particular relationship with the Fourier Transform of a Boolean Function {\em when that Boolean Function is Monotone}. 
In summary our main contributions include:
\begin{enumerate}
\setlength\itemsep{-0.1em}
    \item We provide a precise (and abstract) definition of the MaxCon problem in terms of finding the maximum upper zero of a Monotone Boolean Function defined over the Boolean Cube where, vertices correspond to subsets of data and the output of the MBF correspond to whether the subset can be feasibly covered by a model with set tolerance. 
    \item We link the concept of influences in a MBF to the concept of outlier in MaxCon and provide theoretical analysis to show  that  influences of points belonging to the largest structure in data would be  smallest  under ``ideal''  conditions (defined in Section \ref{sec:method}).
    \item Based on the above analysis, we derive a greedy algorithm that search for the maximum upper zero of a MBF to efficiently solve the MaxCon problem.
\end{enumerate}

\section{Background}
\label{sec:background}
\subsection{Boolean Functions and Influences}
\label{sec:Fourier}



For any Boolean function $f: \left \{ 0, 1\right \}^n \rightarrow \left \{0, 1 \right \}$, the probability that the $i$'th coordinate in a random length-$n$ binary vector, $\mathbf{x}$, ``affects'' the output of the function is defined as the influence \cite{o_donnell_2014}:
\begin{equation}
    \mathrm{Inf}_i\left[ f \right] = \underset{\mathbf{x} \sim \left \{ 0,1 \right \}^n}{\mathrm{\mathbf{Pr}}}\left [ f\left ( \mathbf{x}   \right ) \neq f\left ( \mathbf{x}^{\oplus i}   \right ) \right ]
\end{equation}
where $\mathbf{x}^{\oplus i} $ is equal to $\mathbf{x}$  with the $i$'th coordinate flipped. 
For a MBF, the influence of the $i$'th coordinate is equal to the degree-1 Fourier coefficient \cite{o_donnell_2014}. 

The most obvious strategy for influence approximation would be to uniformly sample the cube. 
\begin{equation}
    \mathrm{Inf}_i\left[ f \right] = \underset{\mathbf{x} \sim \left \{ 0,1 \right \}^n}{\mathbb{E}}  \delta \left [f\left ( \mathbf{x}   \right ) \neq f\left ( \mathbf{x}^{\oplus i}   \right )  \right ]
    \label{equ:uniformInfluence}
\end{equation}
where $\delta \left [\cdot \right ] = 1$ if the condition inside is true (0 otherwise).
However, in Section \ref{sec:inf_Accuracy}, we empirically show that this may not be the most efficient when it comes to solving MaxCon. A better approach would be to use the notion of weighted influence \cite{dinur2005hardness}. In such an approach one typically defines a Bernoulli measure $\mu_q(\mathbf{x})$ over the vertices of the Boolean Cube. Operationally, this can correspond to sampling by selecting to ``turn bits $i$'' independently and with probability $q$. Uniform sampling corresponds to $q = 0.5$. Sampling with low $q$ concentrates the measure towards the bottom of the cube (small Hadamard norm) and high $q$ towards the top. The influences under $\mu_q$ distribution can now be defined as:
\begin{equation}
    \mathrm{Inf}_i^{(q)}\left[ f \right] = \underset{\mathbf{x} \sim \mu_q(\mathbf{x}) }{\mathbb{E}}  \delta \left [f\left ( \mathbf{x}   \right ) \neq f\left ( \mathbf{x}^{\oplus i}   \right )  \right ].
    \label{equ:influence}
\end{equation}
The influences under different $q$ values are analysed further in Section \ref{sec:inf_Accuracy}. 

\subsection{Determining feasibility and the concept of a basis}

Searching for the upper zeros of a MBF will require the evaluation of the function on a given subset. In the case of MaxCon, this involves finding if a given subset is feasible or infeasible. 
The feasibility/infeasibility of a subset can be obtained efficiently via solving the following minmax problem and checking if the resulting objective value is within $\epsilon$ \cite{1544828, tj_2015}: 
\begin{equation}
    \underset{\theta}{\min}~ \underset{i}{\max}~ r_{\mathbf{p}_i} \left( \theta \right)
    \label{equ:minmax}
\end{equation}
The above can be solved exactly for cases where $r_{\mathbf{p}_i} \left( \theta \right)$ is quasiconvex using algorithms based on bisection \cite{eppstein2005quasiconvex}. The solution of equation (\ref{equ:minmax}) for a subset $\mathcal{I}^{(j)}$, will return three quantities: the optimal objective value $g^{(j)}$, basis $\mathcal{B}^{(j)}$ and the fitted model $\theta^{(j)}$. A basis for a set, is a small subset such that the value of the final objective on that subset is equal to the final objective on the whole set \cite{chin2017maximum}. Intuitively, a basis in the MaxCon setting is a subset of points ``that prevents the enclosing structure from shrinking further''. The cardinality of  the basis in MaxCon is connected to the concept of ``combinatorial dimension'' and we denote it by $p+1$ ($p$ is the minimum number of data points required for the unique determination of the model).

\subsection{Related work in computer vision}

{Consensus maximisation is widely used for robust fitting in computer vision. Early work mostly focused on solving the consensus maximisation using randomised hypothesise-and-test techniques \cite{Fischler_1981, loransac, raguram2012usac}. More recently, some focus has shifted towards finding globally optimal solutions \cite{tj_2015, 5995640, Cai_2019, li2009consensus, campbell2017globally, li2007practical}. As MaxCon is known to be NP-hard, the run time of the global methods scale exponentially in the general case \cite{chin2018robust}.  This has lead to the search of deterministic and/or near optimal algorithms \cite{cai2018deterministic, probst2019convex, le2017exact, yang2019quaternion}.}

To the best of our knowledge, ours is the first work in computer vision that explicitly poses the MaxCon problem in terms of finding the maximum upper zero of a MBF. However, there are several works that are related, in some sense, to the proposed framework. 
Though not expressed that way in the original works, the $\textrm{A}^*$ ``tree searches'' of \cite{Cai_2019} and \cite{tj_2015} are in fact searches on this cube (where nodes reached by different paths become {\em repeated nodes} in the tree constructed by starting from the top of the Hasse Diagram). In this context, it is interesting to note that RANSAC (and it's many derivatives) search amongst subsets towards the bottom of the Hasse Diagram (minimal sized subsets or $p$-sized); and use these to ``index'' up to higher subsets by greedily including all data points the fit within tolerance of the model found on the $p$-subset.
Furthermore, the methods that use the $\textrm{L}_\infty$-Norm (equation \ref{equ:minmax}) for outlier removal in the context of geometric fitting, can also be seen as traversing the Boolean cube \cite{sim2006removing, olsson2008polynomial, li2007practical}. However, the above methods do not explicitly utilize the theory and tools associated with MBF.


\section{Proposed Method}
\label{sec:method}

\begin{figure}
    \centering
    \includegraphics[width=0.45\textwidth]{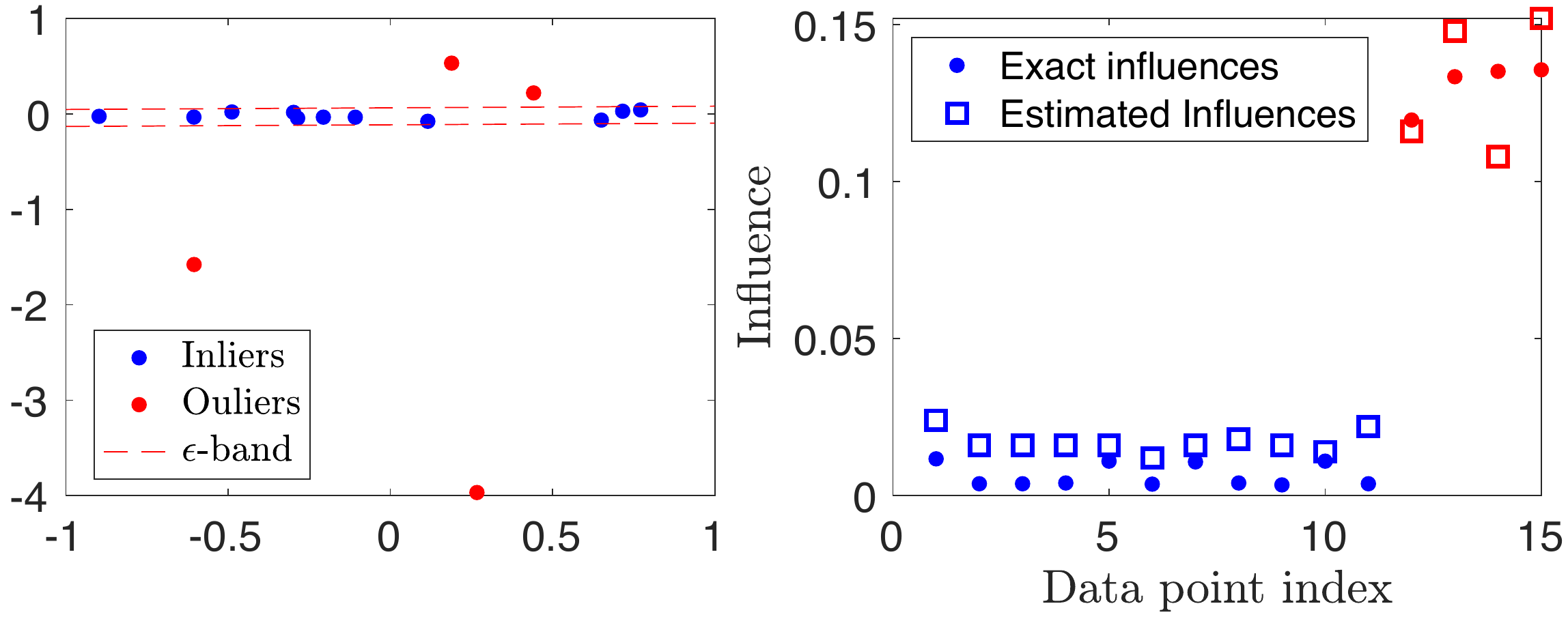}
    \caption{{An example 2D line-fitting problem and the corresponding influences. The data generation procedure is described in detail in Section \ref{sec:inf_Accuracy}. Influences are estimated using equation (\ref{equ:influence}) with $m=1000$ and $q=0.5$. Data points are sorted such that inliers comes before outliers.} }
    \label{fig:influenceCalcExample}
\end{figure}

As described in the introduction, MaxCon can be seen as finding the \textit{maximum upper zero} of a MBF. This paper is devoted to the exploitation of information embedded in the Influence function to find the maximum upper zero, efficiently. To explain the  method,  we  first  note  that  the  influence  of  an inlier  data  point  is  likely  to  be  smaller  than  the  influence of an outlier data point. {An example of this relationship for a  2D line-fitting problem is shown in Figure \ref{fig:influenceCalcExample}}. The essential and intuitive reason is  that  inclusion  of  an  outlier,  into  a  feasible  subset,  most likely turns that subset infeasible (thereby ``influencing'' the function).  In  contrast  adding  an  inlier  leaves  the  set feasible.  
It  is the ``breaking'' or ``creation'' of an infeasible $p+1$ sized basis that is responsible for flipping the outcome of the MBF and therefore  the  addition/deletion  of  an  outlier  triggers  this more than that of an inlier. Below, we prove that the above property of influences hold under ``ideal'' conditions. The proofs of theorems in the following text are available in the supplementary materials.

\subsection{Ideal single structure case}
We first formally define an ideal structure in data. Let $L_k := \left \{ \mathbf{x} \in \left \{ 0,1 \right \}^n : \left \| \mathbf{x} \right \|_1 = k \right \}$ be the level $k$ in the $n$-dimensional Boolean cube and, $L_{\leq k}$ the levels below $k+1$.  
\theoremstyle{definition}
\begin{definition}
Given a monotone Boolean function $f$, for $\mathbf{x}^k \in L_k$ ($p < k \leq n$), $f$ is called ideal with respect to $\mathbf{x}^k$ if
\begin{equation}
    f\left ( \mathbf{x} \right ) = \begin{cases}
0 & \forall \mathbf{x} \in B_{\mathbf{x}^k} \cup L_{\leq p}\\ 
1 & \text{others} 
\end{cases}
\end{equation}
where $B_{\mathbf{x}^k} = \left \{ \mathbf{x} \in \left \{ 0,1 \right \}^n :  d \left (\mathbf{x}, \mathbf{x}^k  \right ) =l ,  \mathbf{x} \in L_{k-l}\right \}
$ for all $0\leq l \leq k-p-1$, is the Boolean sub-cube determined by $\mathbf{x}^k$. Here, $d \left (\cdot, \cdot  \right )$ is the hamming distance.
\end{definition}
Intuitively, an ideal structure has only one structure: points are either inlier or outliers to that single structure, and no other structure exists other than the trivial with $p$ or less points. 

\begin{theorem}
If a monotone Boolean function $f$ is ideal with respect to $\mathbf{x}^k \in L_k$, then
\begin{equation}
    \mathrm{Inf}_i\left[ f \right] = \begin{cases}
\frac{C^{n-1}_p - C^{k-1}_p }{2^n} & \text{if}~i~\text{is inlier} \\ 
\frac{C^{n-1}_p + \sum_{l=p+1}^k C^{k}_l }{2^n} & \text{if}~i~\text{is outlier}
\end{cases}
\end{equation}
\end{theorem}

\begin{corollary}
\begin{equation}
    \mathrm{Inf}_j\left[ f \right] - \mathrm{Inf}_i\left[ f \right] = \frac{1}{2^n} \sum_{l=p+1}^k C_l^k + C_p^{k-1} > 0
\end{equation}
where $i$ is any inlier and $j$ is any outlier.
\end{corollary}
The above shows that for an ideal structure case the influence of an inlier is strictly smaller than the influence of an outlier. 

\subsection{Ideal $K$-structure case}

In what follows, we generalize the above result to ideal $K$-structure ($K \geq 1$), namely there are $K$ upper zeros only and the Boolean sub-cubes determined by these upper zeros are disjoint above level $p$.

\theoremstyle{definition}
\begin{definition}
Let $f$ be a monotone Boolean function and $\left \{ \mathbf{x}^{k_r} \right \}_{r=1}^K$ are upper zeros, where $p < k_1 \leq k_2 \leq \cdots \leq k_K \leq n$, then $f$ is called $K$-ideal with respect to $\left \{ \mathbf{x}^{k_r} \right \}_{r=1}^K$ if
\begin{equation}
    \begin{matrix}
1) & d\left ( B_{\mathbf{x}^{k_i}} \setminus L_{\leq p}, B_{\mathbf{x}^{k_j}} \setminus L_{\leq p} \right )>0 \quad \forall k_i \neq k_j &\\ 
2) & f \left ( \mathbf{x} \right )=\begin{cases}
0 & \forall \mathbf{x} \in \bigcup_{r=1}^K B_{\mathbf{x}^{k_r}} \cup L_{\leq p}  \\ 
1 & \text{others} 
\end{cases}
\end{matrix}
\end{equation}
\label{def:kstruct}
\end{definition}

The first condition states that the sub-cubes determined by $\mathbf{x}^{k_r}$ have no elements in common. 
Let $\mathcal{S}_{\mathbf{x}^{k_r}}^c$ be the set of inlier (if $c=1$) or the set of outliers (if $c=0$) to $\mathbf{x}^{k_r}$. Then we can define 
\begin{equation}
   \mathcal{S}_{c_1c_2 \cdots c_K} = \bigcap_{r=1}^K \mathcal{S}_{\mathbf{x}^{k_r}}^{c_r}  \quad c_r \in \left \{ 0,1\right \}
\end{equation}
which represent the index set of inlier to structures where bit string $c_1c_2 \cdots c_K$ is one. For example, $\mathcal{S}_{11 \cdots 1}$ is the index set of points that are inliers with respect to all $\mathbf{x}^{k_r}$, and $\mathcal{S}_{00 \cdots 0}$ is the index set of points that are outliers with respect to all $\mathbf{x}^{k_r}$.

\begin{theorem}

\begin{equation}
 2^n \mathrm{Inf}_{\mathcal{S}_{c_1\cdots c_K}}\left[ f \right]  = C_p^{n-1} + \sum_{\underset{1\leq r \leq K}{c_r=0}} \sum_{l=p+1}^{k_r} C_l^{k_r} -  \sum_{\underset{1\leq r \leq K}{c_r=1}} C_p^{k_r-1}
\end{equation}
where $\mathrm{Inf}_{\mathcal{S}_{c_1 \cdots c_K}}\left[ f \right]$ denote the influence $\mathrm{Inf}_{i}\left[ f \right]$ of $i \in \mathcal{S}_{c_1 \cdots c_K}$.
\end{theorem}

\begin{corollary}
The influences have the following ordered relationship
\begin{equation}
    \forall \alpha, \beta \in \left \{ 0,1 \right \}^K, \alpha > \beta \Rightarrow \mathrm{Inf}_{\mathcal{S}_{\alpha}}\left[ f \right] < \mathrm{Inf}_{\mathcal{S}_{\beta}}\left[ f \right]
\end{equation}
which means $\mathrm{Inf}_{x}\left[ f \right]$ is a real-valued monotone decreasing Boolean function. If any $ \mathcal{S}_\bullet = \emptyset$ then $\mathrm{Inf}_{\mathcal{S}_\bullet}\left[ f \right]$ is not defined. 
\label{coro:kstruct}
\end{corollary}

Following Corollary \ref{coro:kstruct} and the Definition \ref{def:kstruct} we can see that the influence of a point that belongs to a structure with a higher upper zero is smaller that that of a point belongs to a structure with a lower upper zero. 

\subsection{Finding the maximum upper zero}
\label{sec:method_explanation}
The above theoretical analysis shows that influences of points belonging to the largest structure in data would be smallest under ``ideal'' conditions. Thus, in the ideal case, a very simple process of estimation of the influences, followed by thresholding, would immediately lead to the sought after MaxCon solution. However,  we need to recognise that:
\begin{itemize}
\setlength\itemsep{0em}
    \item The non-ideal case (which applies to all realistic data sets) will be a perturbation away from this, where the influences of inlier and outlier come closer together.
    \item We can only work with estimated influences, not the actual influences. Thus we need to assume that the estimated influences largely follow the ordering given mathematically in our derivations and according with the above intuition.
\end{itemize}
Nonetheless, continuity of behaviour arguments suggest the departure from ideal will be only partial for many data sets; and that relatively standard ways for attacking such scenarios, should still be viable. We empirically verify that this is often true
in our experiments.

Using the above intuition (``\textit{influences of points belonging to the largest structure in data would be smallest}'') we formulate an algorithm - MBF-MaxCon (Algorithm \ref{alg:BMF-maxcon-linf}) that starts with the set of all data points and then gradually removes one data point at a time (data point with the largest influence) until the remaining set of points is within the $\epsilon$ band. In the algorithm, we also use the notion that: \textit{If there are outliers in a subset of data, then at least one of them should belong to the basis returned by solving equation} (\ref{equ:minmax}) \cite{sim2006removing}. This additional constraint enables us to compute only $p+1$ influences per elimination of a data point (rather than for all points in subset) which leads to a more efficient algorithm. 

The estimation of Influence introduces some noise to the proposed algorithms, and the solution returned by them may not include all the inlier points of a given structure. To partially overcome this, we introduce a local expansion step (Algorithm \ref{alg:BMF-maxcon-lexp}). In this step, starting from the initial solution, at each iteration, the distance-1 upper neighborhood (Hasse Diagram) of the current solution is explored and the current solution is updated if there is any feasible set. This process is repeated until there are no feasible subsets in the distance-1 upper neighborhood. This local update guarantees that the algorithm will find an upper zero (local optimal).

\begin{algorithm}[!ht]                      
	\caption{MBF-MaxCon - algorithm for finding the maximum consensus set using influences of BMFs.}          
	\label{alg:BMF-maxcon-linf}                           
	\begin{algorithmic} [1]
		\REQUIRE $\left \{\mathbf{p}_i \right\}_{i=1}^{n}$, $\epsilon$, $m$, $q$.

		\STATE $\mathbf{x} \gets \left [ 1, \dots, 1 \right ]_{\left [ 1 \times n \right ]}$
		
		\REPEAT
		\STATE $\mathcal{I}^{(t)} \gets \left \{ i : x_i = 1 \right \}$
		\STATE Solve equation (\ref{equ:minmax}) for subset $\mathcal{I}^{(t)}$ and obtain $\mathcal{B}^{(t)}$
        
        \STATE Estimate $\mathrm{Inf}_i^{(q)}\left[ f \right] \quad \forall i \in \mathcal{B}^{(t)}$ using equation (\ref{equ:influence})\footnotemark.
        \STATE $e \gets \underset{i \in \mathcal{B}^{(t)} }{\textrm{argmax}}~ \mathrm{Inf}_i^{(q)}\left[ f \right] $
        \STATE $x_e \gets 0$
        
		\UNTIL{$f(\mathbf{x}) = 0$}
		\STATE $\mathbf{x} \gets$ Run local expansion step in (Algorithm \ref{alg:BMF-maxcon-lexp}) 
		\RETURN $\mathcal{I} \gets \left \{ i : x_i = 1 \right \}$
		
	\end{algorithmic}
\end{algorithm}
\footnotetext{When estimating influences, one can use the monotonic nature of $f$ to save some computations  (In a MBF, the function value before the bit flip has some information regarding the value after). The algorithm used for estimating the influences is available in supplementary materials.}

\begin{algorithm}[!ht]                      
	\caption{Local expansion step.}          
	\label{alg:BMF-maxcon-lexp}     
	\begin{algorithmic} [1]
		\REQUIRE $\left \{\mathbf{p}_i \right\}_{i=1}^{n}$, $m$, initial feasible set $\mathbf{x}$.

		\REPEAT
		\STATE updated $\gets$ false
		\FORALL{$i : x_i = 0$}
		\STATE $\bar{\mathbf{x}} \gets \mathbf{x}$;~$\bar{{x}}_i \gets 1$
		\STATE \algorithmicif $~f(\bar{\mathbf{x}}) = 0$ \algorithmicthen $~\mathbf{x} \gets \bar{\mathbf{x}}$; updated $\gets$ true;~ break;
		\ENDFOR
		\UNTIL{updated=false}
		\RETURN $\mathbf{x}$
		
	\end{algorithmic}
\end{algorithm}

It is important to note that one needs to re-estimate the influences at each iteration, $t$ of algorithm \ref{alg:BMF-maxcon-linf}. {If $\mathrm{Inf}_i^{[t]}\left[ f \right] $ is the influence of point $i$ at iteration $t$ of the algorithm, then  $\mathrm{Inf}_i^{[t]}\left[ f \right]  \neq \mathrm{Inf}_i^{[t-1]}\left[ f \right] $.  This is because function at level $t$ is a restricted version of the function at level $t-1$ and the relationship between influences at different levels is derived in supplementary materials.}
Another allied intuition is that though, as mentioned before, noise (from both the estimation process and from the departure of the data from that of being ideal) will raise the level of the influence of some inliers (and decrease the values of some outliers) to the point where the estimated influences of some inliers will be above those of some outliers: at each stage we only remove the {\em largest} influence data point which will be away from the ``polluted'' data division; and that re-estimation afterwards allows the possibility for the re-estimated influences to be ``cleaner''.   

\section{Results}
We evaluated the performance of the proposed algorithms, on both synthetic and real data experiments, and compared those with the state-of-the-art techniques. All experiments were executed in MATLAB on a computer with Intel Xeon E5-1650 CPU, 16GB RAM and Ubuntu 16.04 OS. The publicly available codes were used to obtain the results for improved $\textrm{A}^*$ tree search\footnote{\href{https://github.com/ZhipengCai/MaxConTreeSearch}{Code: MaxConTreeSearch}} ($\textrm{A}^\ast$-NAPA-DIBP) \cite{Cai_2019} and Lo-RANSAC\footnote{\href{https://github.com/ZhipengCai/Demo---Deterministic-consensus-maximization-with-biconvex-programming}{Code: Consensus-maximization-with-biconvex-programming}} \cite{loransac}. Our implementation is publicly available at \href{https://github.com/}{Link removed to preserve anonymity}. 

\subsection{Influence estimation}
\label{sec:inf_Accuracy}

Estimating the influences via equation (\ref{equ:influence}) requires two hyper-parameters: The number of randomly sampled bit-vectors $m$ and, the sampling probability $q$ of the Bernoulli measure $\mu_q(x)$. In this section we explore the effects of the two hyper-parameters on influence estimation using a 2D-line fitting problem. Here, the number of data points ($n$) is set to $15$, out of which 25\% would be outliers ($n_o$). The number of points is chosen to be relatively small as the exact influence calculation time increases exponentially with $n$. A subset of ($n - n_{o}$) randomly selected points (inliers) were then perturbed with uniformly distributed noise in the range $\left [ -0.1, 0.1 \right ]$. The remaining $n_{o}$ data points (outliers) were then perturbed with uniformly distributed noise from $\left [-5, -0.1\right ) \cup \left ( 0.1, 5 \right]$. The inlier threshold $\epsilon$ was set to $0.1$ for all the experiments in this section.  

First, we compare the estimated influences with the exact influences for different combinations of $m$ and $q$. The exact influences, $\overline{\mathrm{Inf}}_i\left[ f \right] $, are computed by taking the expectation in equation (\ref{equ:uniformInfluence}) over all $2^n$ vertices of the Boolean cube. The estimation error of all influence values can then be calculated as:
    $\frac{1}{n} \sum_{i \in \left [ n \right]}  \left ({\mathrm{Inf}}_i^{(q)}\left[ f \right] - \overline{\mathrm{Inf}}_i\left[ f \right]   \right )^2$.

\begin{figure}
    \centering
    \includegraphics[width=0.35\textwidth]{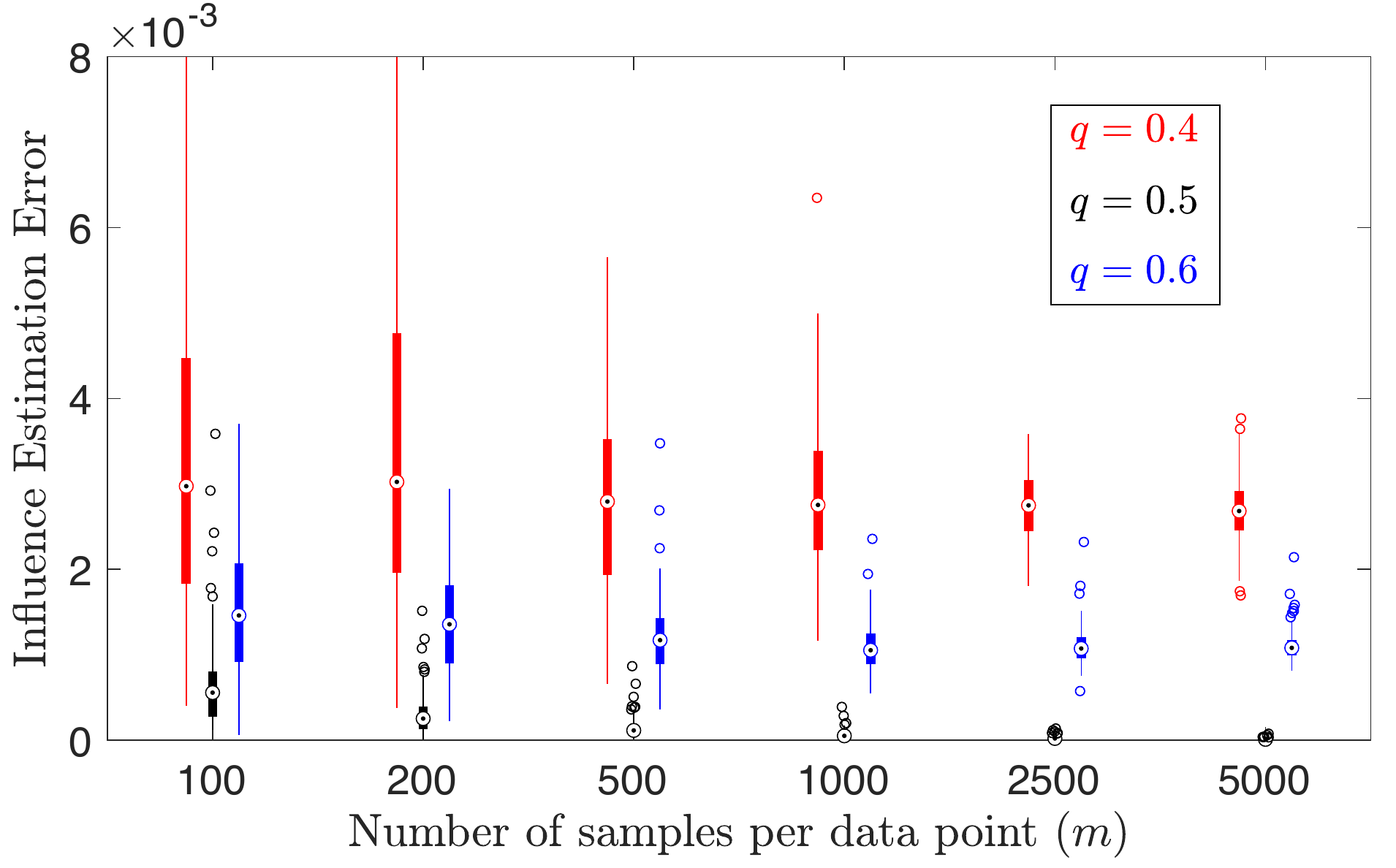}
    \caption{The variation of the influence estimation error with the number of samples used ($m$) and the sampling probability ($q$). The figure show the statistics across 100 random runs for each combination.}
    \label{fig:influenceCalcError}
\end{figure}

The results in Figure \ref{fig:influenceCalcError} shows that, for all $q$ values, the influence estimation error is high when only a few samples are used. However, the error decreases exponentially with increasing the number of samples. We can also see that the estimated influences are closest to their exact values when $q=0.5$. When $q$ is varied in either direction, the error increases. This is because at $q=0.5$, the computed values are an unbiased estimate of the influences. Changing the sampling distribution changes the definition of the orthonormal basis and hence introduces a bias \cite{dinur2005hardness}. 

\begin{figure}[!ht]
     \centering
     \begin{subfigure}[b]{0.225\textwidth}
         \centering
         \includegraphics[width=\textwidth]{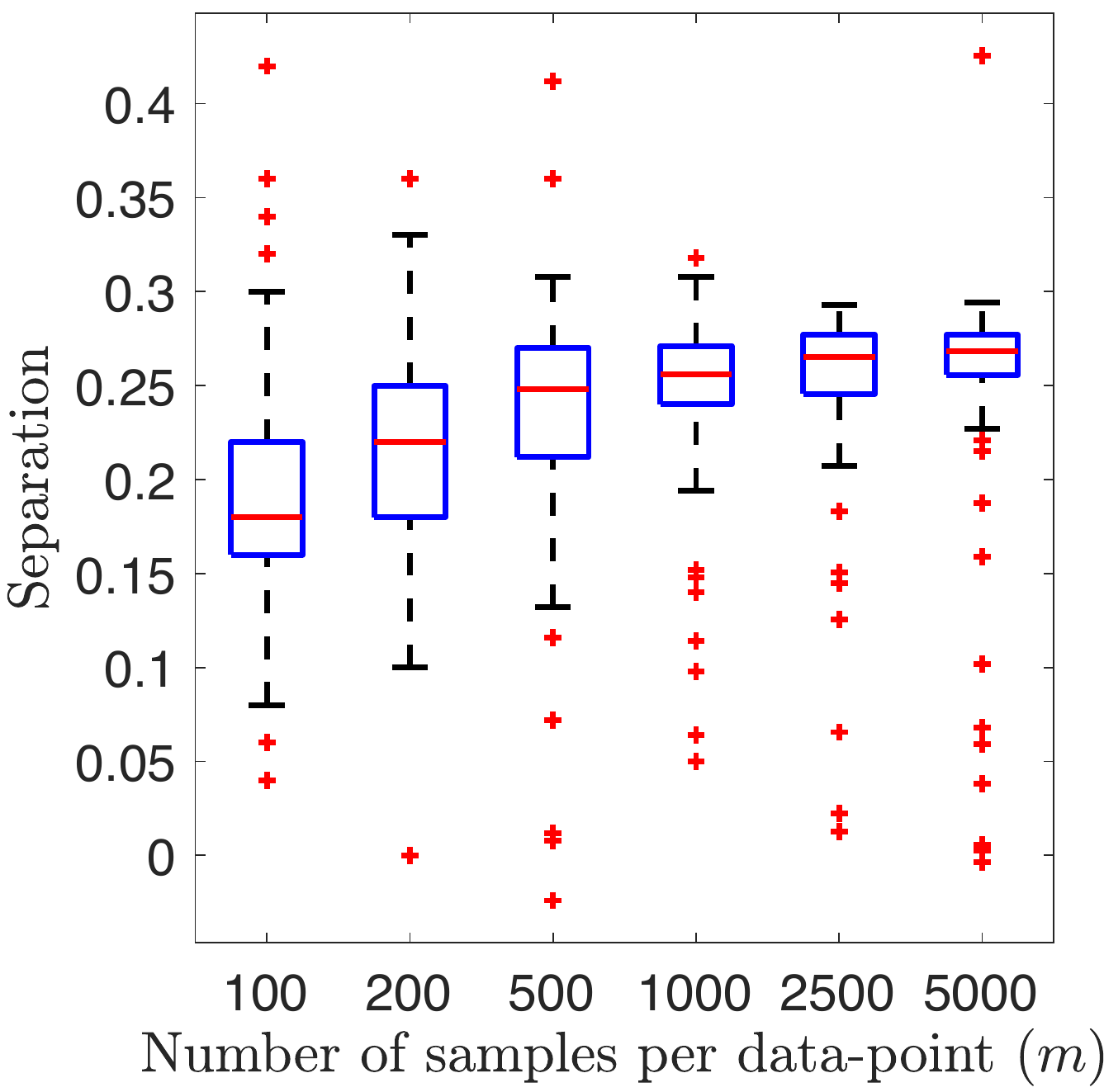} 
         \label{fig:influenceSeparation_subfig1}
     \end{subfigure}
     \begin{subfigure}[b]{0.225\textwidth}
         \centering
         \includegraphics[width=\textwidth]{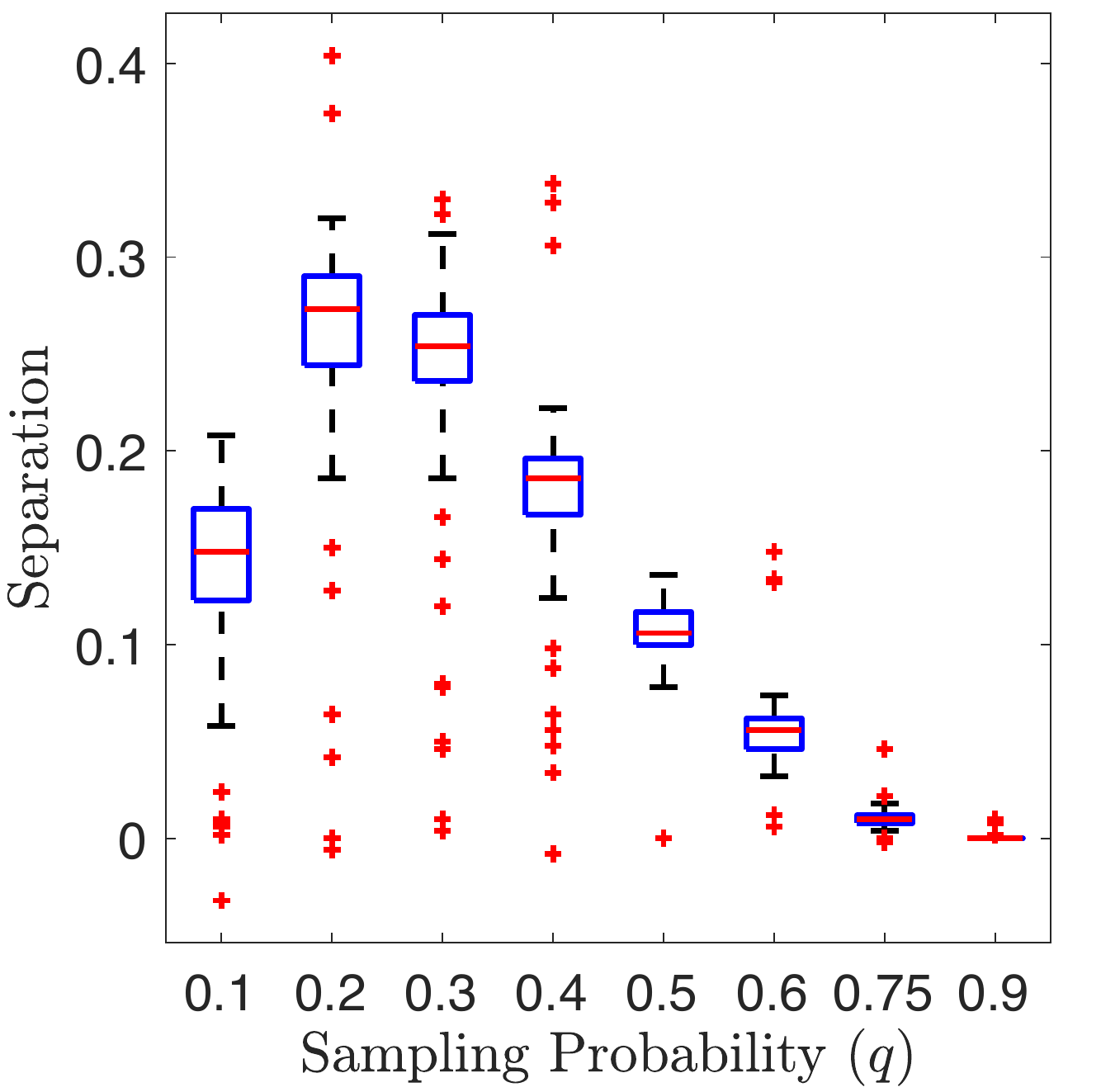} 
         \label{fig:influenceSeparation_subfig2}
     \end{subfigure}
     \vspace{-.2cm}
        \caption{The variation in the separation (between influences of inliers and outliers) with the number of samples used ($m$) and the sampling probability ($q$). The figure show the statistics across 100 random runs for each combination.}
        \label{fig:influenceSeparation}
        \vspace{-.4cm}
\end{figure}

However, in solving MaxCon we do not seek for an unbiased estimate of the influences. What we are after is a definition of influences where the separation between influences of inliers and ourliers are maximum. We used the same synthetic experiment to analyse the separation. Here we define the separation as:
\begin{equation}
 \textrm{Separation} = \underset{i \in \mathcal{D}_{out}}{\min}~ \mathrm{Inf}_i^{(q)}\left [ f \right ] - \underset{j \in \mathcal{D}_{in}}{\min}~ \mathrm{Inf}_j^{(q)}\left [ f \right ]
\end{equation}
where $\mathcal{D}_{in}$ is the set of inlier data points and $\mathcal{D}_{out} = \mathcal{D} \setminus \mathcal{D}_{in}$. The results in Figure \ref{fig:influenceSeparation} show that the separation increase with $m$. However, in relation to $q$, the separation is maximum when the $q$ value is small. Using $q < (p+1)/n$ will also decrease the separation as most samples at this probability will be trivially feasible. In summary, changing $q$ leads to different measures of influence where the order (influences of inliers are smaller than outliers) is much the same, though the separation can vary. In our experiments, we use a $q$ value between $(p+1)/n$ and $0.3$.

\subsection{Controlled experiments with synthetic data}
To study the behaviour of the proposed algorithm under a controlled setting, similar to \cite{Cai_2019}, we conducted experiments on an 8-dimensional robust linear regression problem with synthetically generated data. First, a set of $n=200$ data points on a randomly instantiated model $\theta \in \mathbb{R}^8$ was generated. 
The data set was then perturbed using the same process described in Section \ref{sec:inf_Accuracy} to get a data set that is corrupted by noise and outliers.
In our experiments, the number of outliers, $n_{o}$, were varied in the range of $\left[5, 40\right]$ (upper bound determined by computation time of $\textrm{A}^\ast$-NAPA-DIBP used for obtaining ground truth inliers/outliers). 
The error of a method is computed by comparing the cardinality of the MaxCon solution found by that method ($\left | \mathcal{I}_\bullet \right |$) to the ground truth cardinality obtained using $\textrm{A}^\ast$-NAPA-DIBP ($\left | \mathcal{I}_{\mathrm{A}^\ast} \right |$).

\noindent
\textbf{Ablation study:}
To identify the importance of each component in our overall algorithm, we conduct an ablation study using the above data. Here, we analyse four variants of our algorithm:
    1) \textit{MBF-MaxCon-nR}: Simple algorithm where all the influences are computed at start and the data point with largest influence is removed iteratively until the remaining subset is feasible (no re-estimation of influences at each iteration).
    2) \textit{MBF-MaxCon-nB}: Same as MBF-MaxCon-nR but the influences of all remaining points are recomputed (not just points in basis) at the end of each iteration. 
    3)  \textit{MBF-MaxCon-nL}: Same as algorithm \ref{alg:BMF-maxcon-linf} without the local expansion in line 9.
    4) \textit{MBF-MaxCon}: Proposed algorithm \ref{alg:BMF-maxcon-linf}.

\begin{figure}[!ht]
     \centering
     \begin{subfigure}[b]{0.225\textwidth}
         \centering
         \includegraphics[width=\textwidth]{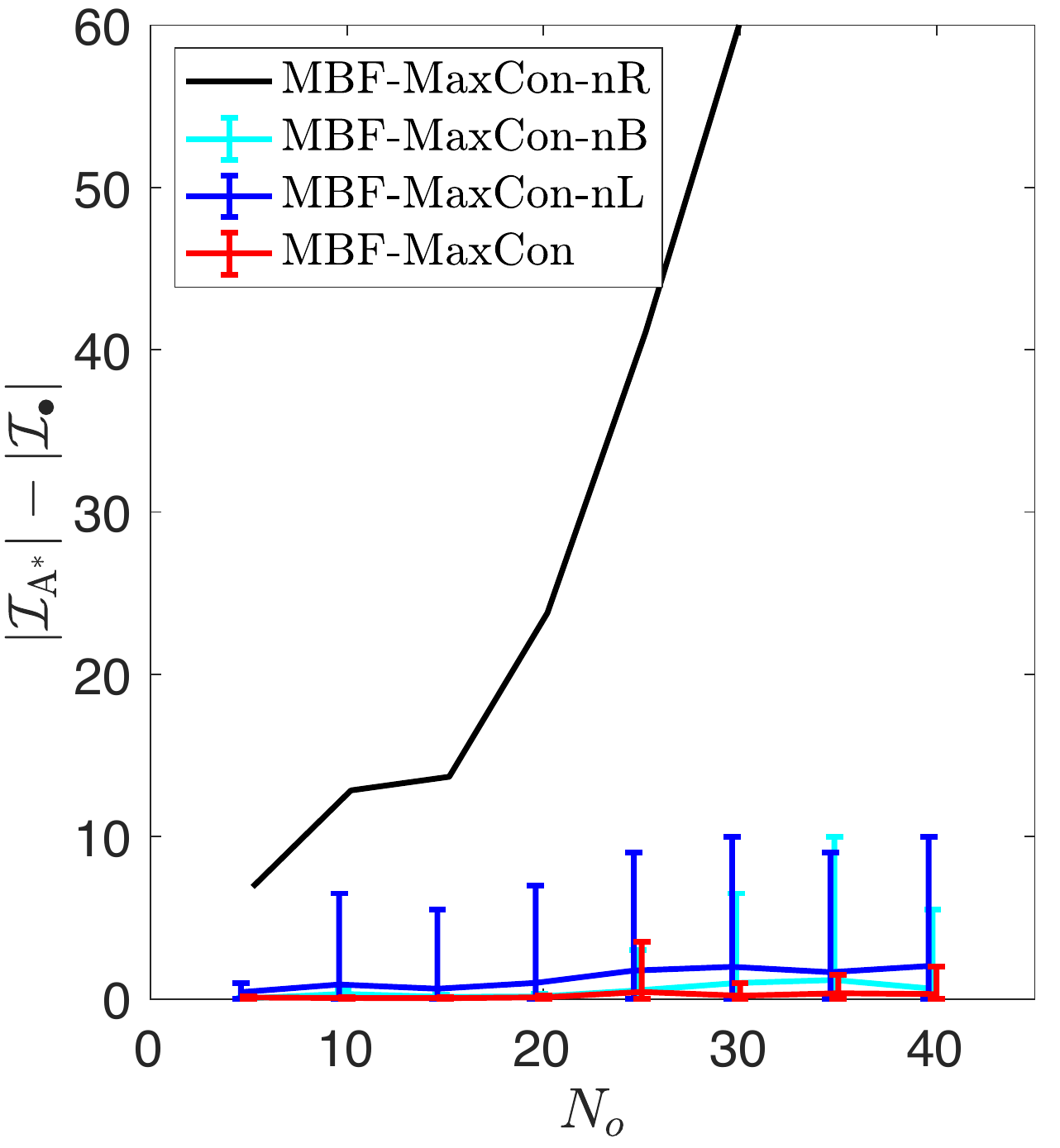} 
         \caption{}
         \label{fig:ablation_subfig1}
     \end{subfigure}
     \begin{subfigure}[b]{0.225\textwidth}
         \centering
         \includegraphics[width=\textwidth]{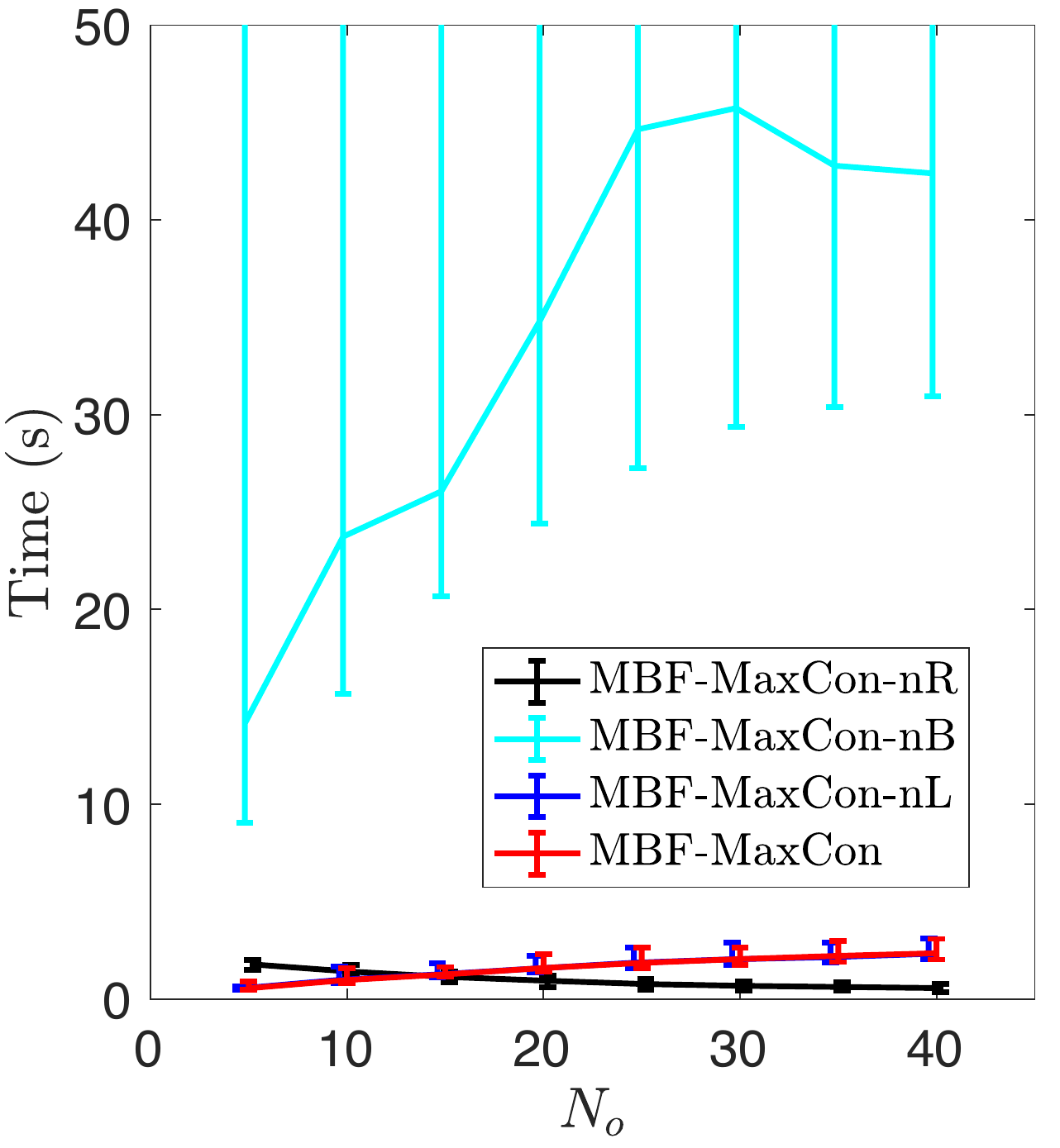} 
         \caption{}
         \label{fig:ablation_subfig2}
     \end{subfigure}
        \caption{Results of the ablation study for 8-dimensional robust linear regression with synthetic data (a) Number of inliers found compared with the global optimal (obtained using $A^*$) and (b) Variation of computational time with number of outliers. The experiments were repeated 100 times and the error-bars indicate the $0.05$-th and $0.95$-th percentile.}
        \label{fig:ablation}
\end{figure}

The error of each variant and the computation times are shown in Fig. \ref{fig:ablation}. The results show that re-estimating influences at the end of each iteration has a significant effect on the final outcome (see section \ref{sec:method_explanation} for explanation). Furthermore, the results show that, on average,  the use of both local expansion and basis has helped to move the solutions closer to the global optimal. The results also show that the most significant contribution in terms of the use of basis is in computational efficiently.

\noindent
\textbf{Comparative analysis:}
Next, we compare the performance of the proposed method with the most relevant methods in literature:  $\textrm{A}^\ast$-NAPA-DIBP \cite{Cai_2019}, RANSAC \cite{Fischler_1981}
and Lo-RANSAC \cite{loransac}. Both RANSAC variants were run with the number of RANSAC iterations set to match the computation time of MBF-MaxCon. 

\begin{figure}[!ht]
     \centering
     \begin{subfigure}[b]{0.225\textwidth}
         \centering
         \includegraphics[width=\textwidth]{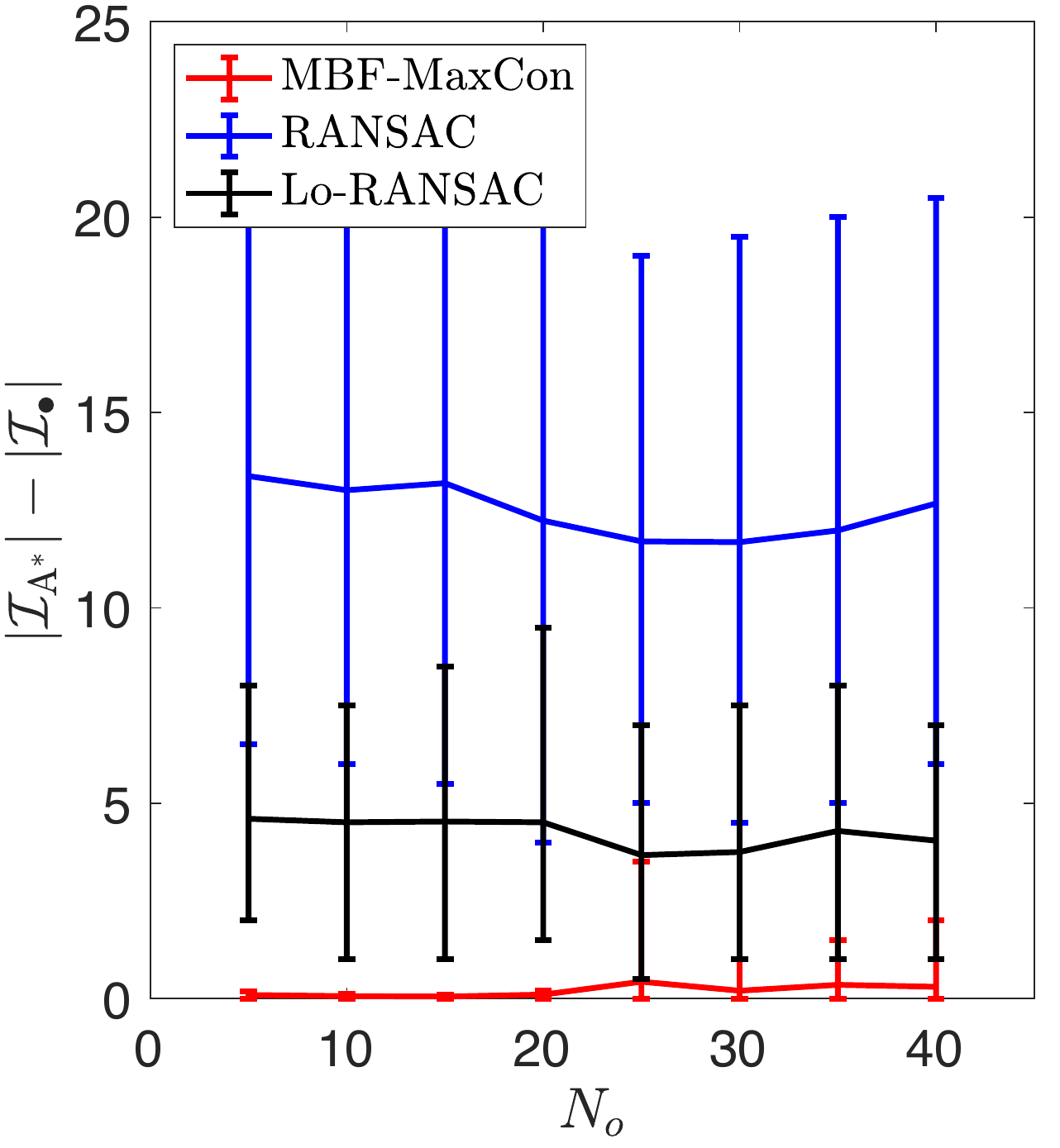} 
        \caption{}
         \label{fig:syntheticData_example_subfig1}
     \end{subfigure}
     \begin{subfigure}[b]{0.225\textwidth}
         \centering
         \includegraphics[width=\textwidth]{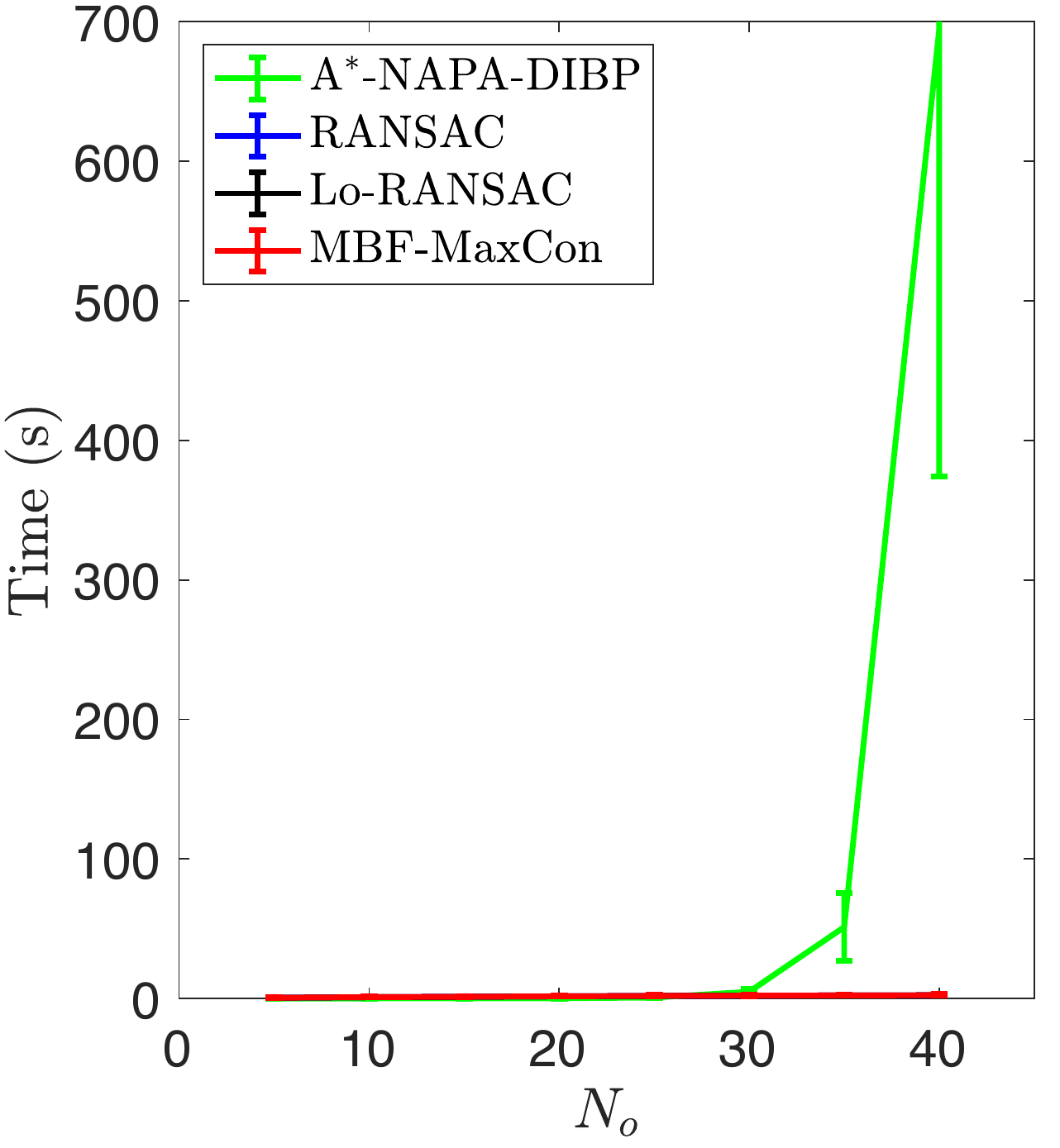} 
            \caption{}
         \label{fig:syntheticData_example_subfig2}
     \end{subfigure}
        \caption{Results for 8 dimensional robust linear regression with synthetic data (a) Number of inliers found compared with the global optimal (obtained using $A^*$) and (b) Variation of computational time with number of outliers. The experiments were repeated 100 times and the error-bars indicate the $0.05$-th and $0.95$-th percentile.}
        \label{fig:syntheticData_example}
\end{figure}

The computation time for the above methods are shown in Figure \ref{fig:syntheticData_example_subfig2}. The figure shows that when the number of outliers are low ($< 30$) $\textrm{A}^\ast$-NAPA-DIBP converges to a solution relatively quickly. However, the computational time of $\textrm{A}^\ast$-NAPA-DIBP increases exponentially with the number of outliers. On the other hand, the computational time of the proposed algorithms increase linearly with the number of outliers. This is clearly predictable as our algorithms take one step across each level and the deeper down is the MaxCon solution, proportionally longer is the ``search''. $\textrm{A}^\ast$-NAPA-DIBP has a much more sophisticated search that allows backtracking of routes explored and this causes the exponential behaviour when that is heavily exercised. Figure \ref{fig:syntheticData_example_subfig1} shows the difference between the number of inliers returned by $\textrm{A}^\ast$-NAPA-DIBP and other methods. On average the proposed method MBF-MaxCon returns a solution with usually close to the same number of inliers as the A* method. The figure also shows the 0.05th and 0.95th percentile distances from $A^*$ solution over 100 random runs. This shows that in few cases the solution returned by MBF-MaxCon can be up to 2-4 inliers away from the optimal solution (around 1\% error). The main summary is that the proposed methods never ``go exponential'', unlike $A^*$, in runtime, but compares often favourably in terms of accuracy.

\subsection{Linearized fundamental matrix estimation}
In this section we examine the performances of the proposed methods for linearized fundamental matrix estimation. Provided that the point matches between two views are given as $[\textbf{p}_1, \textbf{p}_2]$ where $\textbf{p}_j = (x_j, y_j, 1)^\top$ is a coordinate of a point in view $j$, each rigid motion in the scene can be modeled using the fundamental matrix $F \in \mathcal{R}^{3 \times 3}$  as \cite{Torr1997}: $\textbf{p}_1^\top F \textbf{p}_2 = 0$. In our experiments we use the linearized version presented in \cite{tj_2015} together with the algebraic error \cite{hartley2003multiple} (chapter 11).  For each image pair, the input is a set of SIFT \cite{lowe1999object} feature matches generated using VLFeat \cite{vedaldi08vlfeat}.

\noindent
\textbf{Single dominant motion:}
Following \cite{Cai_2019}, we used the first five crossroads image pairs from the sequence ``00'' of the KITTI Odometry dataset \cite{geiger2012we} in our experiments.  The inlier threshold $\epsilon$ is set to $0.03$ for all image pairs. The Number of inliers returned by each method ($n_i$) and the computation times are shown in the first part of Table \ref{tab:lin_fund_res}. The results reported for the probabilistic methods are the mean (min, max) over 100 random runs. The results show that the proposed methods on average have produced solutions that are close to the optimum solution returned by $\textrm{A}^\ast$-NAPA-DIBP \cite{Cai_2019}.

\begin{figure}
    \centering
    \includegraphics[width=0.35\textwidth]{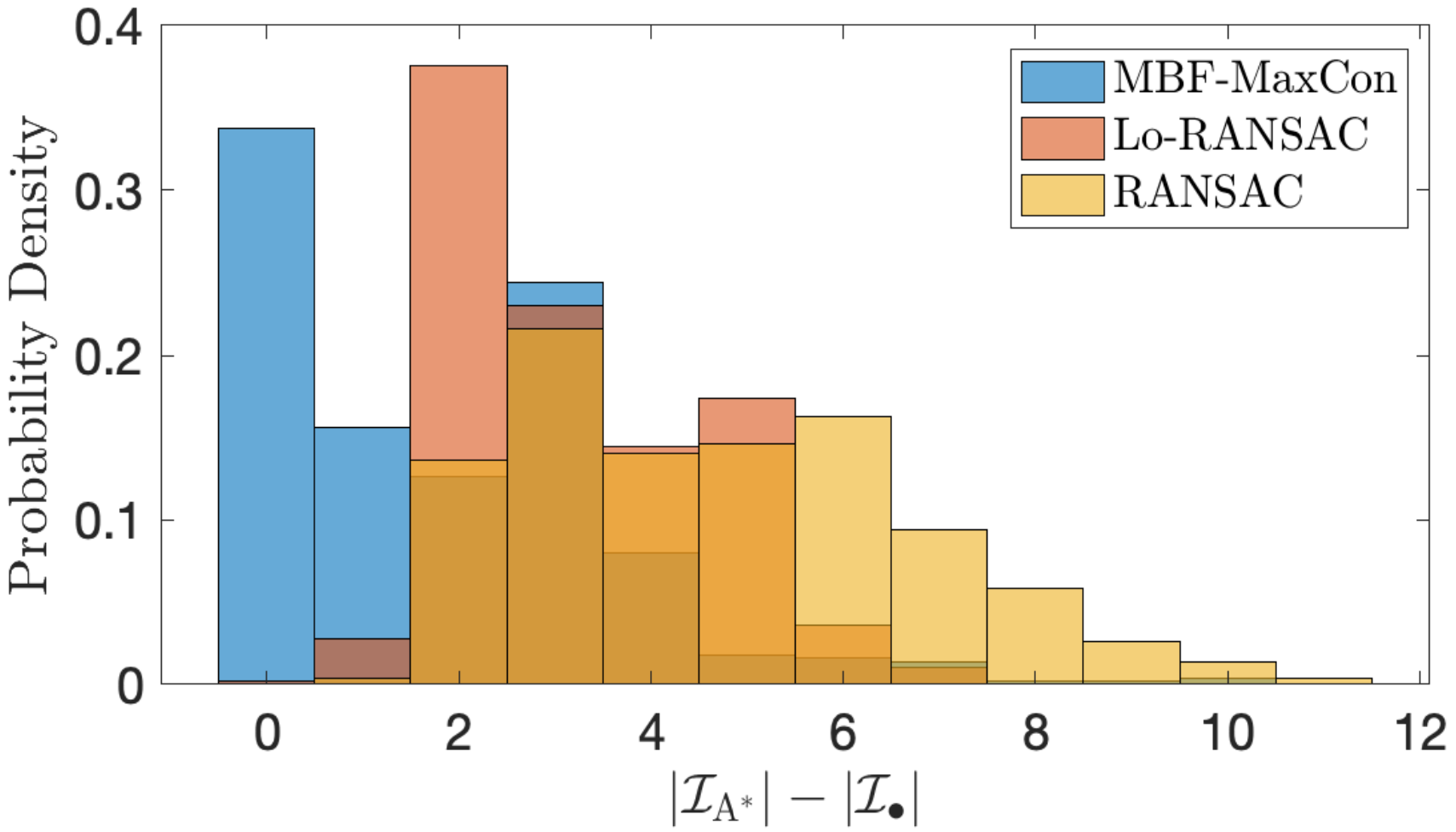}
    \caption{The distribution of errors by each algorithm over 100 repeated runs for all the frames in sequence “00” of the KITTI Odometry data set.}
    \label{fig:linearizedfund_hist}
    \vspace{-.4cm}
\end{figure}

The distribution of errors by each algorithm over 100 repeated runs for all the frames in sequence “00” of the KITTI Odometry data set is shown in Figure \ref{fig:linearizedfund_hist}. The main message is that we operate in a time cost regime a little better than $\textrm{A}^\ast$-NAPA-DIBP and around the same as we allowed for Lo-RANSAC but we generally get much closer to $\textrm{A}^\ast$-NAPA-DIBP performance - including often finding the optimal, which RANSAC or Lo-RANSAC rarely do in this experiment.

\begin{table*}[]
\footnotesize
\centering
\caption{Linearized fundamental matrix estimation result. ``Same Comp.'' refers to running RANSAC with the same time budget as MBF-MaxCon and, ``sp=0.99'' refers to running RANSAC with the success probability $0.99$.  }
\renewcommand{\arraystretch}{0.8}
\begin{tabular}{llcccccc}
\hline
\hline
\multicolumn{1}{l}{}                &          & \textbf{\begin{tabular}[c]{@{}c@{}}A*-NAPA\\      -DIBP\end{tabular}} & \textbf{MBF-MaxCon}                                               & \textbf{\begin{tabular}[c]{@{}c@{}}RANSAC\\      Same Comp.\end{tabular}} & \textbf{\begin{tabular}[c]{@{}c@{}}Lo-RANSAC\\      Same Comp.\end{tabular}} & \textbf{\begin{tabular}[c]{@{}c@{}}RANSAC\\      sp=0.99\end{tabular}} & \textbf{\begin{tabular}[c]{@{}c@{}}Lo-RANSAC\\      sp=0.99\end{tabular}} \\ \hline 
\multirow{2}{*}{\textbf{104-108}}   & $n_i$     & 289                                                                   & \begin{tabular}[c]{@{}c@{}}288.52\\       (289, 285)\end{tabular} & \begin{tabular}[c]{@{}c@{}}282.03\\      (286, 277)\end{tabular}          & \begin{tabular}[c]{@{}c@{}}284.54\\      (287, 283)\end{tabular}             & \begin{tabular}[c]{@{}c@{}}271.18\\      (282, 254)\end{tabular}      & \begin{tabular}[c]{@{}c@{}}281.41\\      (284, 276)\end{tabular}         \\
                                    & Time (s) & 10.96                                                                 & 1.78                                                              & 1.78                                                                      & 1.78                                                                         & 0.004                                                                 & 0.04                                                                     \\ \hline
\multirow{2}{*}{\textbf{198-201}}   & $n_i$     & 296                                                                   & \begin{tabular}[c]{@{}c@{}}293.10\\       (296, 291)\end{tabular} & \begin{tabular}[c]{@{}c@{}}291.88\\      (294, 290)\end{tabular}          & \begin{tabular}[c]{@{}c@{}}292.87\\       (294, 291)\end{tabular}            & \begin{tabular}[c]{@{}c@{}}287.00\\      (293, 272)\end{tabular}      & \begin{tabular}[c]{@{}c@{}}290.35\\       (293, 287)\end{tabular}        \\
                                    & Time (s) & 4.04                                                                  & 2.05                                                              & 2.05                                                                      & 2.05                                                                         & 0.004                                                                 & 0.04                                                                     \\ \hline
\multirow{2}{*}{\textbf{417-420}}   & $n_i$     & 366                                                                   & \begin{tabular}[c]{@{}c@{}}364.44\\       (366, 359)\end{tabular} & \begin{tabular}[c]{@{}c@{}}363.54\\      (365, 361)\end{tabular}          & \begin{tabular}[c]{@{}c@{}}364.02\\       (365, 363)\end{tabular}            & \begin{tabular}[c]{@{}c@{}}357.38\\      (364, 343)\end{tabular}      & \begin{tabular}[c]{@{}c@{}}362.33\\       (364, 359)\end{tabular}        \\
                                    & Time (s) & 7.93                                                                  & 2.59                                                              & 2.59                                                                      & 2.59                                                                         & 0.004                                                                 & 0.06                                                                     \\ \hline
\multirow{2}{*}{\textbf{579-582}}   & $n_i$     & 523                                                                   & \begin{tabular}[c]{@{}c@{}}520.68\\       (523, 514)\end{tabular} & \begin{tabular}[c]{@{}c@{}}518.04\\      (521, 511)\end{tabular}          & \begin{tabular}[c]{@{}c@{}}520.89\\       (522, 520)\end{tabular}            & \begin{tabular}[c]{@{}c@{}}502.31\\      (519, 463)\end{tabular}      & \begin{tabular}[c]{@{}c@{}}517.30\\       (512, 497)\end{tabular}        \\
                                    & Time (s) & 4.28                                                                  & 3.22                                                              & 3.22                                                                      & 3.22                                                                         & 0.004                                                                 & 0.11                                                                     \\ \hline
\multirow{2}{*}{\textbf{738-742}}   & $n_i$     & 462                                                                   & \begin{tabular}[c]{@{}c@{}}460.97\\       (462, 457)\end{tabular} & \begin{tabular}[c]{@{}c@{}}455.02\\      (459, 451)\end{tabular}          & \begin{tabular}[c]{@{}c@{}}457.24\\       (460, 455)\end{tabular}            & \begin{tabular}[c]{@{}c@{}}438.55\\      (455, 409)\end{tabular}      & \begin{tabular}[c]{@{}c@{}}451.35\\       (459, 435)\end{tabular}        \\
                                    & Time (s) & 2.97                                                                  & 2.72                                                              & 2.72                                                                      & 2.72                                                                         & 0.005                                                                 & 0.1                                                                      \\ \hline \hline
\multirow{2}{*}{\textbf{breadcube}} & $n_i$     & $\sim$                                                                & \begin{tabular}[c]{@{}c@{}}64.36\\      (68, 58)\end{tabular}     & \begin{tabular}[c]{@{}c@{}}61.77\\      (65, 59)\end{tabular}             & \begin{tabular}[c]{@{}c@{}}63.66\\      (66, 61)\end{tabular}                & \begin{tabular}[c]{@{}c@{}}59.04\\      (65, 55)\end{tabular}         & \begin{tabular}[c]{@{}c@{}}62.17\\      (66, 59)\end{tabular}            \\
                                    & Time (s) & \textgreater{}3600                                                    & 19.07                                                             & 19.07                                                                     & 19.07                                                                        & 0.993                                                                 & 1.07                                                                     \\ \hline
\multirow{2}{*}{\textbf{breadtoy}}  & $n_i$     & $\sim$                                                                & \begin{tabular}[c]{@{}c@{}}107.48\\      (115, 102)\end{tabular}  & \begin{tabular}[c]{@{}c@{}}105.24\\      (111, 102)\end{tabular}          & \begin{tabular}[c]{@{}c@{}}107.23\\      (111, 104)\end{tabular}             & \begin{tabular}[c]{@{}c@{}}100.31\\      (107, 93)\end{tabular}       & \begin{tabular}[c]{@{}c@{}}105.21\\      (109, 101)\end{tabular}         \\
                                    & Time (s) & \textgreater{}3600                                                    & 38.37                                                             & 38.37                                                                     & 38.37                                                                        & 0.526                                                                 & 0.68                                                                     \\ \hline
\multirow{2}{*}{\textbf{cubetoy}}   & $n_i$     & $\sim$                                                                & \begin{tabular}[c]{@{}c@{}}58.51\\      (61, 52)\end{tabular}     & \begin{tabular}[c]{@{}c@{}}54.54\\      (58, 52)\end{tabular}             & \begin{tabular}[c]{@{}c@{}}56.14\\      (58, 55)\end{tabular}                & \begin{tabular}[c]{@{}c@{}}52.20\\      (56, 48)\end{tabular}         & \begin{tabular}[c]{@{}c@{}}55.24\\      (57, 52)\end{tabular}            \\
                                    & Time (s) & \textgreater{}3600                                                    & 15.38                                                             & 15.38                                                                     & 15.38                                                                        & 0.45                                                                  & 0.65                                                                     \\ \hline \hline
\end{tabular}
\label{tab:lin_fund_res}
\end{table*}

\noindent
\textbf{Multiple motions:} 
The above data set contains a single dominant motion and the number of outliers are limited (around 13-22). To study the behaviour of the proposed algorithm in the presence of multiple structures, we conducted experiments on three sequences from the AdelaideRMF data-set \cite{wong2011dynamic}. The inlier threshold $\epsilon$ was set to $0.015$ for all image pairs. In these sequences, there are two rigidly moving objects. The results in the second part of Table \ref{tab:lin_fund_res} show that the method $\textrm{A}^\ast$-NAPA-DIBP did not find a solution after $3600$s (1 hour), where as the proposed methods found solutions in around 15-40 seconds. Once again, when compared to Lo-RANSAC, we generally obtain higher consensus sets (and thus implicitly closer to what $A^*$ is capable of, but on these data sets, would require astronomically more computation.

\section{Conclusion}
We have applied a new perspective to the long standing problem of MaxCon. This perspective recognises that the underlying mathematical object is a Monotone Boolean Infeasibility function, defined over the Boolean Cube. Such a perspective immediately identifies a rich mathematical theory that can be applied. Very probably, we have only scratched that surface here. But we have been able to take at least one element of the theory (i.e. Influence) 
and link that concept to the concept of outlier (in MaxCon) and shown that already, without borrowing further from the rich theory, that we can derive algorithms that are already at least competitive, in some aspects. Specifically: 
\begin{enumerate}
\setlength\itemsep{0em}
\item The approach sometimes achieves the true MaxCon, whereas RANSAC rarely achieves the MaxCon solution. Indeed, it is well known that this is a feature of RANSAC based methods in general. 
\item The approach can (mostly) achieve close to $A^*$ (provably optimal) in a similar time budget or faster.
\item Unlike $A^*$ the approach will never go exponential in runtime or memory requirements, and - as above - unlike RANSAC variants, it can more reliably obtain the optimal or close to optimal result - given a similar time budget.
\end{enumerate}

The proposed algorithm can be used on model fitting problems where the residuals, $r_{\mathbf{p}_1} \left( \theta \right)$, are quasi-convex and the run-time of the algorithm would depend on the existence of the an efficient oracle to evaluate the feasibility/infeasibility of a subset of points.  


{\small
\bibliographystyle{ieee_fullname}
\bibliography{egbib}
}

\end{document}